\pdfoutput=1

\documentclass[11pt]{article}

\usepackage{acl}

\usepackage{lipsum}
\usepackage{times}
\usepackage{latexsym}
\usepackage{subcaption}
\usepackage{multirow}
\usepackage{xcolor}
\usepackage[T1]{fontenc}
\usepackage{setspace}
\usepackage{multicol}
\usepackage{tikz}

\usepackage[utf8]{inputenc}

\usepackage{booktabs}
\usepackage{tabularx}
\usepackage{graphicx}
\usepackage{subcaption}
\usepackage{microtype}

\usepackage{inconsolata}
\usepackage[most]{tcolorbox}
\usepackage{authblk}

\definecolor{green}{RGB}{51,102,0}
\definecolor{red}{RGB}{204, 0, 0}
\usepackage[table]{xcolor}

\title{The Role of Model Confidence on Bias Effects in Measured Uncertainties for Vision-Language Models}

\author{Xinyi Liu$^{1}$ $\quad$ Weiguang Wang$^{2}$ $\quad$ Hangfeng He$^{1}$\\
$^{1}$University of Rochester $\quad$ $^{2}$Johns Hopkins University\\
{\tt\small xinyi.liu1@simon.rochester.edu, www@jhu.edu}\\
{\tt\small hangfeng.he@rochester.edu}}

\begin{document}

\maketitle
\begin{abstract}

With the growing adoption of Large Language Models (LLMs) for open-ended tasks, accurately assessing epistemic uncertainty, which reflects a model’s lack of knowledge, has become crucial to ensuring reliable outcomes. However, quantifying epistemic uncertainty in such tasks is challenging due to the presence of aleatoric uncertainty, which arises from multiple valid answers. While bias can introduce noise into epistemic uncertainty estimation, it may also reduce noise from aleatoric uncertainty. To investigate this trade-off, we conduct experiments on Visual Question Answering (VQA) tasks and find that mitigating prompt-introduced bias improves uncertainty quantification in GPT-4o.
Building on prior work showing that LLMs tend to copy input information when model confidence is low, we further analyze how these prompt biases affect measured epistemic and aleatoric uncertainty across varying bias-free confidence levels with GPT-4o and Qwen2-VL. We find that all considered biases have greater effects in both uncertainties when bias-free model confidence is lower. Moreover, lower bias-free model confidence is associated with greater bias-induced underestimation of epistemic uncertainty, resulting in overconfident estimates, whereas it has no significant effect on the direction of bias effect in aleatoric uncertainty estimation. These distinct effects deepen our understanding of bias mitigation for uncertainty quantification and potentially inform the development of more advanced techniques. \footnote{\url{https://github.com/XinyiLiu0227/Uncertainty_Quantification_Bias}}

\end{abstract}

\section{Introduction}

\begin{figure}
    \centering
    \includegraphics[width=\linewidth]{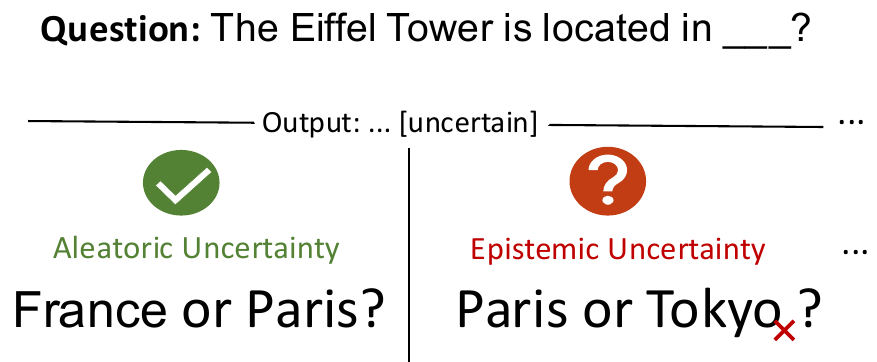}
    \vspace{-6mm}
    \caption{Uncertainty between valid answers (e.g., France and Paris) reflects aleatoric uncertainty, while uncertainty between Paris and Tokyo reflects epistemic uncertainty due to the model’s lack of knowledge.}
    \vspace{-4mm}
    \label{fig:example}
\end{figure}

Robust quantification of Large Language Models' (LLMs) confidence in their answers is vital for trust and safety in critical applications \citep{hendrycks2021unsolved, rudner2024key}. Without effective confidence ranking, accurate predictions may be overlooked, while inaccurate predictions may be prioritized and lead to harmful outcomes \citep{geifman2017selective}. 

Much of the existing literature leverages uncertainty to approximate a model's confidence in its answers \citep{guo2017calibration, malinin2020uncertainty}. Model uncertainty can stem from aleatoric uncertainty, epistemic uncertainty, or both. Importantly, only epistemic uncertainty is indicative of the model's confidence, since it captures the limitations of the model's knowledge. In contrast, aleatoric uncertainty stems from the irreducible randomness of the true answer distribution and persists even if the model has perfect knowledge. As such, the true goal of ``uncertainty quantification'' is to quantify the epistemic uncertainty. When two predictions exhibit similar total uncertainty, the one driven by aleatoric uncertainty indicates a more knowledgeable and confident model than one dominated by epistemic uncertainty. Figure~\ref{fig:example} illustrates this distinction through an example where the model is uncertain for different underlying reasons. 

Traditional uncertainty quantification methods, however, typically estimate total uncertainty. This is because they often operate under the single-answer assumption, where aleatoric uncertainty is absent. Yet in real-world scenarios with multiple valid answers, distinguishing between the two becomes crucial. 


In settings where each question has only one valid answer and uncertainty is thus purely epistemic, it may be intuitive that the presence of bias \citep{ye2024spurious, yang2024identifying, seo2022information}, namely spurious features that models rely on without understanding the true semantic meanings, can lead to inaccurate uncertainty estimation based on biased generation probabilities. Therefore, mitigating bias can improve the effectiveness of uncertainty quantification based on generation probabilities \citep{jiang2023calibrating}. However, the potential presence of aleatoric uncertainty introduces additional complexity. Bias may also reduce aleatoric uncertainty by concentrating probability mass on a single or smaller subset of valid answers. In such cases, bias may reduce the noise introduced by aleatoric uncertainty, potentially facilitating a clearer estimation of epistemic uncertainty.

We investigate whether mitigating prompt-introduced biases can enhance uncertainty quantification with the presence of aleatoric uncertainty,  using GPT-4o, one of the most advanced multimodal LLMs. These biases arise from arbitrary and unavoidable choices in spurious features that do not alter the underlying semantics when using a single prompt, such as phrasing, answer position, verbalizer assignment, and image shape \citep{wang2023large, liu2024empirical, gavrikov2024vision, ye2024mm}. Our results show that bias mitigation consistently enhances uncertainty quantification with the presence of aleatoric uncertainty, without requiring access to the internal model state.  Specifically, removing text-based biases boosts AUROC \citep{hanley1983method, mcdermott2024closer} by approximately 7\%. Motivated by this, we further examine how bias affects epistemic and aleatoric uncertainty separately.

Earlier research predominantly tackles the aleatoric uncertainty from different phrasings of the same semantic meaning, often by semantic equivalence calculations \citep{kuhn2023semantic, farquhar2024detecting, lin2023generating}. Recent work \citep{ahdritz2024distinguishing, yadkori2024believe} has shifted focus towards more general scenarios, where multiple distinct semantic meanings are valid \citep{jiang2022uncertainty, jia2024combining, barandas2024evaluation}. These two studies find that models are more likely to copy information from prompts under high epistemic uncertainty than under high aleatoric uncertainty, which may be interpreted as a form of confirmation bias \citep{nickerson1998confirmation, shi2024argumentative}. Therefore, we hypothesize that the impact of the prompt-introduced biases examined in our earlier experiments on epistemic uncertainty amplifies with lower bias-free model confidence, whereas its impact on aleatoric uncertainty remains relatively insensitive to confidence levels. 


Most multi-label Natural Language Processing datasets were introduced early and are now well studied, allowing LLMs to achieve near-perfect performance with minimal uncertainty \citep{yadkori2024believe}. We therefore construct visual-language datasets where LLM performance is not yet saturated, enabling analysis of both text-based and image-based prompt biases. 


\begin{figure}
    \centering
    \includegraphics[width=0.9\linewidth]{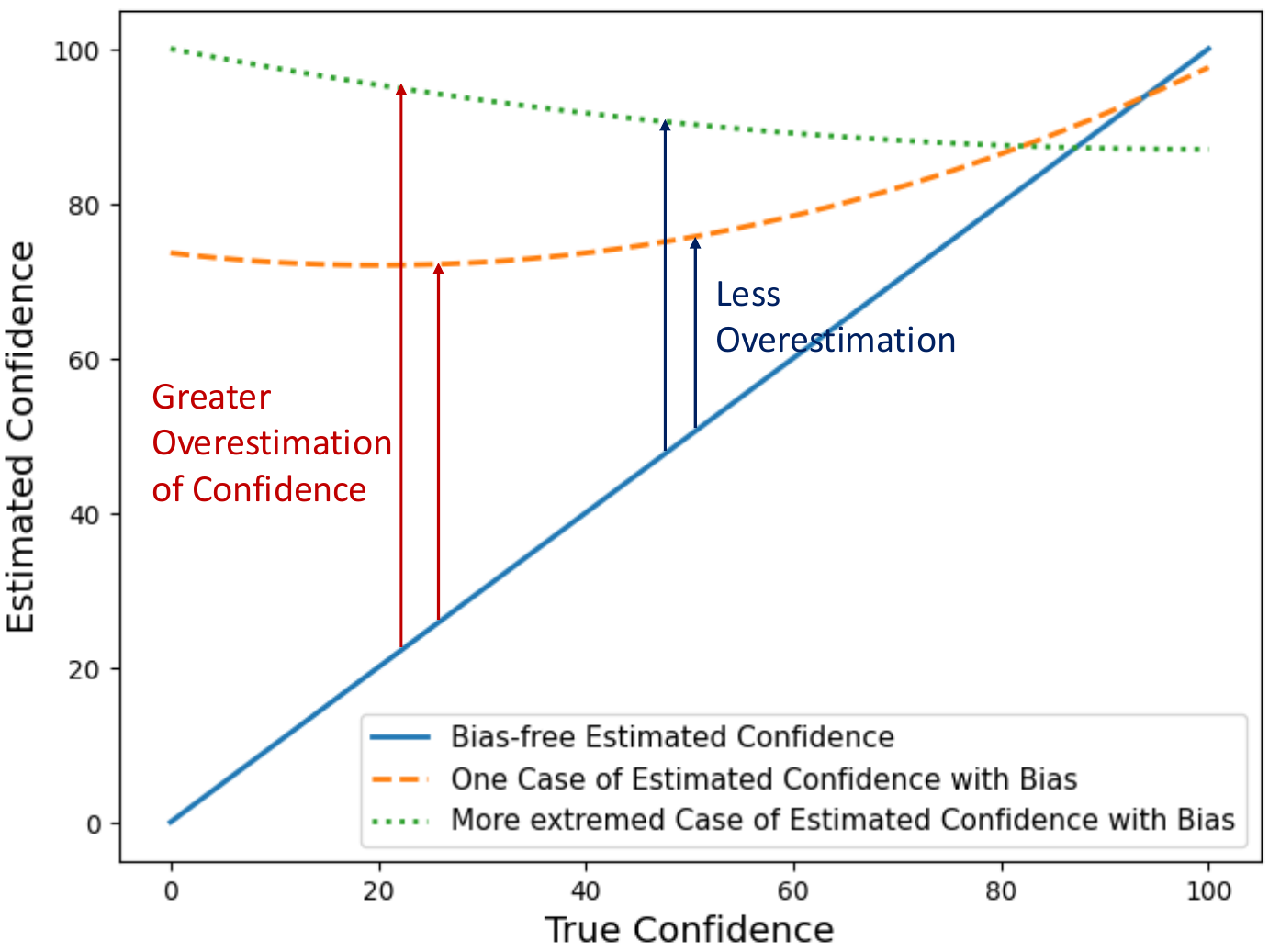}
    \vspace{-2mm}
    \caption{Systematically greater overestimation of confidence in lower-confidence instances can flatten the estimated confidence curve, undermining ranking robustness. Sometimes it even reverses the correct order.}
    \vspace{-4mm}
    \label{fig:overconfidence}
\end{figure}

For both closed-source model GPT-4o \citep{hurst2024gpt} and open-source model Qwen2-VL \citep{wang2024qwen2}, our findings show that lower bias-free model confidence correlates with stronger bias effects, estimated by the absolute change in both epistemic and aleatoric uncertainty measured with and without bias. However, this correlation is notably weaker for aleatoric uncertainty than for epistemic uncertainty. 

As illustrated in Figure \ref{fig:overconfidence}, assigning more inflated confidence scores to lower-confidence instances undermines the robustness of ranking by measured confidence. Notably, this pattern has been observed in human behavior~\citep{sulistyawati2011prediction}. In extreme cases, such distortions may even reverse the correct ranking: when the bias-free confidence in A exceeds that in B, the biased estimates incorrectly favor B over A. To better understand this directional distortion in confidence estimation, we examine how bias interacts with different sources of uncertainty. We find that bias-induced underestimation of epistemic uncertainty is greater when the model is genuinely less confident, resulting in overconfident estimates. In contrast, the bias-induced directional shift in aleatoric uncertainty estimation shows no significant association with confidence.

The distinct effects of bias on epistemic and aleatoric uncertainty deepen our understanding of bias mitigation in uncertainty quantification. This understanding may also guide the development of more advanced methods.

\section{Related Work}

\paragraph{Uncertainty Quantification with a Single Valid Answer.}

Traditional machine learning models treat total uncertainty as a measure of confidence when each question has a single valid answer~\citep{hendrycks2016baseline, lakshminarayanan2017simple, guo2017calibration, wang2022uncertainty}. In single-choice classification problems like MMLU \citep{hendrycks2020measuring}, studies~\citep{rae2021scaling, kadavath2022language} show that LLMs are generally well-calibrated. 

Reinforcement Learning with Human Feedback (RLHF) has complicated uncertainty estimation \citep{ouyang2022training}. Studies \citep{xiong2023can, zhou2024relying} show that RLHF-trained LLMs often overestimate their confidence, raising concerns about the reliability of self-reported uncertainty. Moreover, \citet{huang2023large} and \citet{feng2024don} found that self-reflection alone is insufficient for accurately assessing uncertainty.

\citet{jiang2023calibrating} found that rephrasing and reordering prompts improve uncertainty quantification in single-answer settings. While their approach partially overlaps with ours in textual perturbation, we extend the analysis to multi-answer scenarios that involve aleatoric uncertainty and additional prompt-introduced biases, including image-based biases. Crucially, we further examine how these biases affect the two uncertainties differently across varying confidence levels, offering a deeper understanding of the bias mitigation method.

\paragraph{Uncertainty Quantification with a Single Semantic Valid Answer.}

Prior work on LLM uncertainty with aleatoric components mainly focuses on variability in generating semantically equivalent outputs, using benchmarks such as CoQA~\citep{reddy2019coqa}, TriviaQA~\citep{joshi2017triviaqa}, and AmbigQA~\citep{min2020ambigqa}.


Proposed techniques include training auxiliary classifiers \citep{kamath2020selective, cobbe2021training} and leveraging internal model states \citep{ren2022out, burns2022discovering, lin2023generating}, requiring additional training or model access.
Semantic equivalence has proven to be effective in reducing aleatoric uncertainty from phrasing variability without access to internal model states \citep{kuhn2023semantic, farquhar2024detecting}. Research by \citet{huang2023look} observed that sample-based methods outperform single-inference approaches.

Building on these findings, we shift focus from phrasing variation to the challenge of multiple semantically valid answers, aiming to capture the distinct characteristics of epistemic and aleatoric uncertainty.

\paragraph{Uncertainty Quantification with Multiple Semantic Valid Answers.}
\begin{figure*}
    \centering
    \scalebox{0.85}{
    \includegraphics[width=\linewidth]{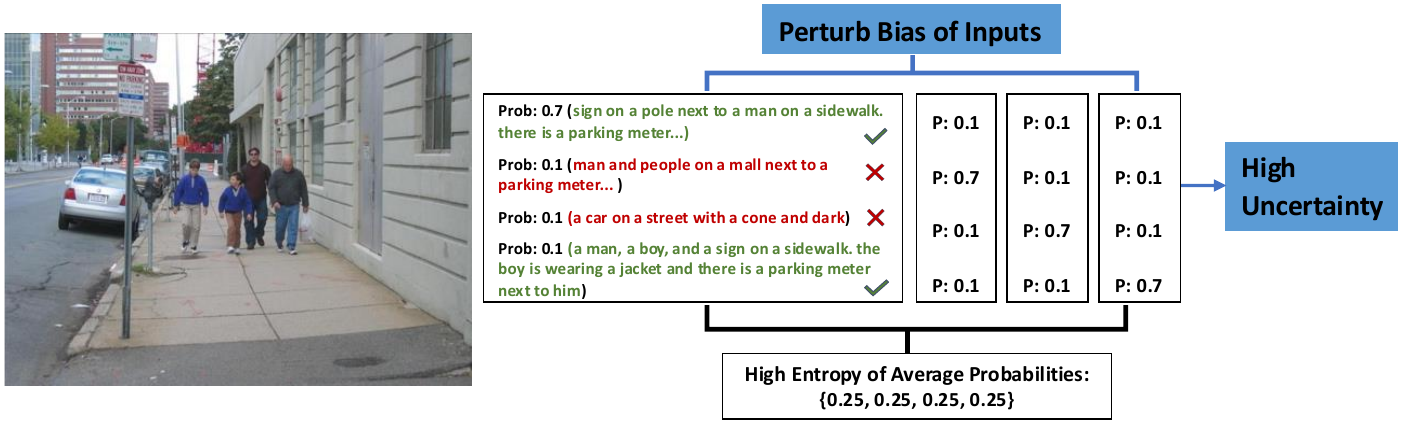}
    }
    \vspace{-2mm}
    \caption{Perturb prompts to shuffle bias factors to estimate bias-free uncertainty.
    }
    \label{fig:teaser}
\vspace{-0.2cm}
\end{figure*}

Uncertainty estimation becomes more complex with multiple semantically valid answers. \citet{ahdritz2024distinguishing} tackled this by assuming larger models capture aleatoric uncertainty, while a smaller model head is trained to predict it. They also observed that LLMs are more likely to copy input information when epistemically uncertain compared to aleatorically uncertain. \citet{yadkori2024believe} built on similar findings by using mutual information to estimate epistemic uncertainty, measuring answer distribution dependency on provided hints through iterative prompting.

This growing body of work underscores the need to distinguish epistemic from aleatoric uncertainty with multiple semantically valid answers. We extend this by analyzing how biases introduced by relying on a single prompt affect these two measured uncertainties across different model confidences. In addition, \citet{yadkori2024believe} preselected multi-label queries with high entropy (> 0.7) from the WordNet dataset \citep{fellbaum1998wordnet}, where LLMs achieve near-perfect performance. This approach results in instances with high total uncertainty but correct outputs, which may not reflect real-world data distributions. We use unfiltered datasets to better capture practical challenges.

\section{The Role of Bias in Uncertainty Quantification}
\label{sec:methodology}

While bias might add noise to epistemic uncertainty estimation, it also may reduce the noise introduced by aleatoric uncertainty. We evaluate this trade-off using GPT-4o, one of the most advanced multimodal LLMs, to assess whether mitigating the prompt-introduced biases improves uncertainty quantification under aleatoric uncertainty.

\citet{ahdritz2024distinguishing} and \citet{yadkori2024believe} both found that LLMs are more likely to copy input information under high epistemic uncertainty but not high aleatoric uncertainty. Inspired by these findings, we further analyze how these prompt-introduced biases impact each type of uncertainty estimation separately, aiming to provide deeper insight.

\subsection{Epistemic and Aleatoric Uncertainty}

Epistemic uncertainty arises from uncertainty in distinguishing correct from incorrect predictions, reflecting the model's lack of knowledge or confidence. In contrast, aleatoric uncertainty stems from uncertainty among multiple valid answers and exists even with perfect world knowledge.

Building on the proven effectiveness of semantic equivalence in addressing phrasing variability, particularly the use of LLM-based Natural Language Inference~\citep{farquhar2024detecting}, we focus on the challenge of multiple valid answers with distinct meanings. To address this, we adopt a multiple-choice format with two semantically distinct correct options and two incorrect ones. This design provides sufficient data for analysis while offering a clear conceptual framework, without introducing additional variance from semantic-equivalence resolution. For generalizing uncertainty quantification from classification to open-ended generation, please refer to Appendix B of \citet{jiang2023calibrating}.

 In uncertainty quantification (see Section~\ref{sec:uncertainty quantification}), ground-truth information is unavailable. However, for analyzing bias impact, we use ground-truth labels to quantify epistemic and aleatoric uncertainty separately. We estimate epistemic and aleatoric uncertainty using epistemic entropy and aleatoric entropy, respectively. We define epistemic entropy as the entropy over the probability of a correct prediction (i.e., the summed probabilities of all valid answers) and the individual probabilities of each incorrect prediction. Let $i$ denote a potential output, and ``correct'' the set of valid answers:

\begingroup
\scriptsize
\begin{align}
    P(\text{correct}) &= \sum_{i \in \text{correct}} P(i) \\
    \text{Epistemic Entropy} = &- P(\text{correct}) \log P(\text{correct}) \nonumber \\
    &- \sum_{i \notin \text{correct}} P(i) \log P(i)
\end{align}
\endgroup
Aleatoric entropy is defined as the entropy over the normalized distribution of correct answers:

\begingroup
\setlength{\abovedisplayskip}{6pt}
\setlength{\belowdisplayskip}{6pt}
\scriptsize
\begin{align}
\text{Aleatoric Entropy} = - \sum_{i \in \text{correct}} 
\frac{P(i)}{P(\text{correct})} \log \frac{P(i)}{P(\text{correct})}
\end{align}
\endgroup
Consequently, the total entropy over the full output distribution, which is commonly used to estimate model uncertainty, can be decomposed into epistemic and aleatoric entropy as follows. A detailed proof is provided in Appendix~\ref{sec:appendix proof}.

\scriptsize
\begin{align}
\label{eq:decompose}
    \text{Entropy} = \text{Epistemic Entropy} + P(\text{correct}) \times \text{Aleatoric Entropy}
\end{align}

\normalsize
\subsection{Prompt-Introduced Biases}
We consider three text-based biases and three image-based biases. The text-based biases include:
\vspace{-0.2cm}
\paragraph{Phrasing Bias.} LLMs often rely on spurious linguistic correlations, making predictions without fully understanding context~\citep{wang2021identifying, si2023spurious}. We mitigate phrasing bias by rephrasing prompts while preserving semantic meaning to average out probability shifts caused by bias.
\vspace{-0.2cm}
\paragraph{Positional Bias.} LLMs are known to exhibit sensitivity to the positions of input options~\citep{wang2023large, liu2024empirical}. We shuffle the positions of the options to neutralize the probability shift from positional bias across prompts.
\vspace{-0.2cm}
\paragraph{Label Bias.} While label bias falls under linguistic features like phrasing bias, shuffling assigned labels offers a more targeted intervention than general paraphrasing. \citet{liu2024empirical} highlighted its significant impact in GPT-3.5 and GPT-4.

Although image-based biases are often reduced through image perturbations during training \citep{shorten2019survey}, we remain interested in exploring whether insights from text-based biases can also be applied to image-based biases. The three image-based biases we consider are:
\vspace{-0.2cm}
\paragraph{Shape Bias.} The shape bias of vision models has been discussed in several studies~\citep{he2023shift, gavrikov2024vision}, where models rely on shape cues to generate their outputs.
\vspace{-0.2cm}
\paragraph{Orientation Bias.} The orientation of images can influence the predictions of vision models, a phenomenon known as orientation bias~\citep{henderson2021biased, ye2024mm}. 
\vspace{-0.2cm}
\paragraph{Low-level Feature Bias.} Injecting noise into images can mitigate biases by reducing reliance on low-level features, such as texture, lighting, and contrast \citep{shorten2019survey}.

More details of prompts perturbation strategies to mitigate biases are provided in Appendix \ref{sec:appendix perturb}. 

\subsection{Uncertainty Quantification in the Presence of Aleatoric Uncertainty}
\label{sec:uncertainty quantification}

We explore bias mitigation for uncertainty quantification, aiming to estimate a model’s confidence in its outputs without ground truth access by reducing prompt-introduced biases, as depicted in Figure \ref{fig:teaser}.

Unlike the mutual information approach proposed by the recent work \citep{yadkori2024believe}, which injects hints into prompts to measure copying behavior, our method operates in a smaller search space by directly targeting biases in default prompts, avoiding broader searches. Specifically, we address both text- and image-based biases unavoidably introduced by a single prompt, as identified in prior work \citep{wang2023large, liu2024empirical, gavrikov2024vision, ye2024mm}.

\subsection{Bias Effects on Measured Uncertainties}
As many top-performing models are closed-source, understanding their behavior as observable without internal states is crucial. We examine how prompt-introduced biases affect measured epistemic and aleatoric uncertainty, offering insights that can be leveraged for both open- and closed-source models.

To assess the impact of bias, we compare entropy values from single prompts to those averaged over multiple bias-shuffled prompts (see Figure~\ref{fig:teaser}). Specifically, we measure: (1) bias effect as the \textit{absolute change} in epistemic and aleatoric entropy, and
(2) bias-induced underestimation as the \textit{decrease} in entropy from the averaged distribution to the single prompt. While the averaged distribution across bias-shuffled prompts may not be entirely bias-free, it is relatively bias-reduced reference~\citep{wang2023large, liu2024empirical} and we refer to it as ``bias-free'' for convenience.

We perform two separate linear regressions to examine the relationship between bias-free confidence levels (independent variable) and each of the two bias effect measures (dependent variable).

\section{Experiments}
\label{sec:experiments}

\begin{table}[!ht]
\centering 
\small
\scalebox{0.85}{
\begin{tcolorbox}

\textbf{Prompt Template}

You are given an image and a set of descriptions. Your task is to evaluate each description and determine whether it is true based on the image.

Below are the descriptions:

\textcolor[rgb]{0,0,0.9}{\{Label\_0\}: \{Option\_0\}}

\textcolor[rgb]{0,0,0.9}{\{Label\_1\}: \{Option\_1\}}

\textcolor[rgb]{0,0,0.9}{\{Label\_2\}: \{Option\_2\}}

\textcolor[rgb]{0,0,0.9}{\{Label\_3\}: \{Option\_3\}}

Provide one index of the descriptions that are true, regardless of the number of descriptions that you believe are true. Return your response as a single index without any additional explanations or text. Here is an example format for your response:

0

Use the provided format and structure for your response.
\end{tcolorbox}
}

\caption{The Vanilla Prompt used to obtain greedy outputs from Large Language Models for evaluating their correctness. An example is provided in Appendix \ref{sec:appendix perturb}.}
\label{tab:prompt}
\vspace{-3mm}
\end{table}

\paragraph{Dataset.} We use the VL\_checklist~\citep{zhao2022vl} and CREPE datasets~\citep{ma2023crepe}, which contain numerous images with human-verified positive and negative descriptions. In contrast, some datasets \citep{thrush2022winoground, tong2024eyes} contain image descriptions but lack multiple correct and incorrect ones per image, while others \citep{ray2023cola, liu2023visual} include only a limited number. We randomly select two correct and two incorrect descriptions and present them in a random order to ensure unbiased LLM evaluation.


These datasets evaluate more advanced model capabilities, compositional reasoning~\citep{hua2024mmcomposition}, compared to early multi-label datasets such as WordNet where current LLMs achieve near-perfect performance. To balance data coverage and budget, we create 1,000 questions from 1,000 images per dataset.

\paragraph{Evaluation Metrics.} 
We adopt the AUROC metric for uncertainty quantification, following prior studies~\citep{band2022benchmarking, kuhn2023semantic, lin2023generating, farquhar2024detecting}. AUROC measures how well confidence scores rank instances with correct versus incorrect predictions and is robust to class imbalance~\citep{mcdermott2024closer}. 

For further analysis, we use linear regression coefficients and p-values to examine how bias-free model confidence influences bias-induced changes in measured epistemic and aleatoric uncertainty. Regression coefficients indicate the direction and magnitude of this relationship: a positive coefficient suggests greater bias effects at higher confidence levels, while a negative coefficient implies that higher confidence reduces bias impact. P-values assess statistical significance, with low values (typically $\leq$ 0.05) indicating a meaningful effect rather than one due to chance.

\paragraph{Models.}
Given the popularity and strong performance of the GPT series, we select the latest stable version of GPT-4o (`gpt-4o-2024-11-20') available at the time. Additionally, we extend our empirical analysis to the open-source LLM Qwen2-VL (`Qwen2-VL-72B-Instruct-GPTQ-Int4'). 

Due to budget constraints, we evaluate only two randomly selected biases for Claude-3.7-Sonnet (`claude-3-5-sonnet-20241022') on CREPE, with the results presented in Appendix \ref{sec: appendix empirical}.

\paragraph{Experimental Settings.}
With OpenAI's closed-source LLMs now providing top-20 token probabilities, we compute prediction probabilities across all options directly for GPT-4o. When token probabilities are not exposed, a common strategy is to approximate them through sampling, which has been shown effective~\citep{farquhar2024detecting}. Following this approach, we approximate the token probabilities for Claude-3.7-Sonnet using 10 sampled prompts. 

Bias-free model confidence is estimated by summing the probabilities of correct options averaged across bias-shuffled prompts. We further extend our experiments by approximating model inconfidence using bias-free epistemic entropy (higher entropy indicates lower confidence), with the results presented in Appendix \ref{sec: appendix empirical}.

Following \citet{kuhn2023semantic} and \citet{farquhar2024detecting}, we approximate greedy decoding by using a single output generated at a very low temperature (1e-15) as the model's `best generation' for assigning correctness labeling, using the prompt shown in Table \ref{tab:prompt}. While closed-source LLMs may still exhibit variation at zero temperature, this approach remains consistent with established research.

\citet{farquhar2024detecting} found that sampling settings, like temperature and top-P, minimally affect sampling-based uncertainty quantification. Based on this, we fix generation parameters (temperature = 0.9, top-P = 1) for sampling from bias-shuffled prompts to ensure consistency and avoid unnecessary tuning. We run ten shuffled prompts for each type of bias, aligning with the sample sizes used in previous sampling-based methods~\citep{huang2023look, kuhn2023semantic, farquhar2024detecting} and the per-iteration sample count in iterative-based methods~\citep{yadkori2024believe}.

\section{Results and Analysis}
\label{sec:results}

\subsection{Uncertainty Quantification Through Bias Mitigation}

\begin{table}[ht]
\centering
{\fontsize{6.6pt}{7.2pt}\selectfont
\scalebox{0.8}{
\begin{tabular}{lccc}
\toprule
\textbf{Methods} & \textbf{\#Inference} & \textbf{VL\_Checklist} & \textbf{CREPE} \\ 
\toprule
Mutual Information & 20 & 0.6782 & 0.5973   \\
\midrule
Repetitive-based \#Answers & 10 & 0.6763 & 0.5821 \\
\cellcolor[HTML]{ECF4FF}Rephrased-based \#Answers (proposed) & \cellcolor[HTML]{ECF4FF}10 & \cellcolor[HTML]{ECF4FF}0.7328 & \cellcolor[HTML]{ECF4FF}0.6106 \\
\midrule
Single-inference Prob & 1 & 0.7349 & 0.5801  \\
Repetitive-based Prob & 10 & 0.7233 & 0.6017 \\
\cellcolor[HTML]{ECF4FF}Rephrase-based Prob (proposed) & \cellcolor[HTML]{ECF4FF}10 & \cellcolor[HTML]{ECF4FF}0.7762 & \cellcolor[HTML]{ECF4FF}0.6513\\
\midrule
Single-inference Entropy & 1 & 0.7492 & 0.5870 \\
Repetitive-based Entropy & 10 & 0.7412 & 0.6084 \\
\cellcolor[HTML]{ECF4FF}Rephrase-based Entropy (proposed)& \cellcolor[HTML]{ECF4FF}10 & \cellcolor[HTML]{ECF4FF}0.7779 & \cellcolor[HTML]{ECF4FF}0.6442\\
\cellcolor[HTML]{ECF4FF}Reorder-based Entropy (proposed)& \cellcolor[HTML]{ECF4FF}10 & \cellcolor[HTML]{ECF4FF}0.7844 & \cellcolor[HTML]{ECF4FF}0.6299 \\
\cellcolor[HTML]{ECF4FF}Relabel-based Entropy (proposed)& \cellcolor[HTML]{ECF4FF}10 & \cellcolor[HTML]{ECF4FF}0.7665 & \cellcolor[HTML]{ECF4FF}0.6406 \\
\cellcolor[HTML]{ECF4FF}Rephrase+Reorder+Relabel-based Entropy (proposed) & \cellcolor[HTML]{ECF4FF}10*3 & \cellcolor[HTML]{ECF4FF}\textbf{0.8123} & \cellcolor[HTML]{ECF4FF}\textbf{0.6588} \\
\cellcolor[HTML]{ECF4FF}Resize-based Entropy (proposed)& \cellcolor[HTML]{ECF4FF}10 & \cellcolor[HTML]{ECF4FF}0.7605 & \cellcolor[HTML]{ECF4FF}0.6219 \\
\cellcolor[HTML]{ECF4FF}Rotate-based Entropy (proposed)& \cellcolor[HTML]{ECF4FF}10 & \cellcolor[HTML]{ECF4FF}0.7565 & \cellcolor[HTML]{ECF4FF}0.6204 \\
\cellcolor[HTML]{ECF4FF}Noise-based Entropy (proposed)& \cellcolor[HTML]{ECF4FF}10 & \cellcolor[HTML]{ECF4FF}0.7535 & \cellcolor[HTML]{ECF4FF}0.6252 \\
\cellcolor[HTML]{ECF4FF}Resize+Rotate+Noise-based Entropy (proposed)& \cellcolor[HTML]{ECF4FF}10*3 & \cellcolor[HTML]{ECF4FF}\textit{0.7699} & \cellcolor[HTML]{ECF4FF}\textit{0.6287} \\
\bottomrule
\end{tabular}
}
}

\caption{This table presents the AUROC scores for uncertainty quantification with GPT-4o. While the Repetitive-based method shows minimal improvement, mitigation of any single bias consistently enhances performance on both datasets. Furthermore, combining methods targeting different biases further improves performance over individual methods.}
\label{tab:model_perforamance_definite}
\vspace{-2mm}
\end{table}

\begin{table*}[h]
    \centering
    \scriptsize
    \begin{tabular}{ccccccccc}
        \toprule
         \multirow{2}{*}{\textbf{Dataset}} & \multirow{2}{*}{\textbf{Bias}} & \multirow{2}{*}{\textbf{Metrics}} & \multicolumn{3}{c}{\textbf{GPT-4o}} & \multicolumn{3}{c}{\textbf{Qwen2-VL}}\\
         \cmidrule(lr){4-6} \cmidrule(lr){7-9}
         & & &\textbf{Epistemic} & \textbf{Aleatoric}  & \textbf{Ratio Epi./Ale.} & \textbf{Epistemic} & \textbf{Aleatoric}  & \textbf{Ratio Epi./Ale.} \\
        \hline
          \multirow{12}{*}{VL\_Checklist} & \multirow{2}{*}{Phrasing} & \scalebox{0.9}{Coefficients} & \textcolor{green}{- 0.2300} & \textcolor{green}{- 0.0579} & \textbf{3.97}  & \textcolor{green}{- 0.0332} & \textcolor{green}{- 0.0123} & \textbf{2.70}   \\
         & & \scalebox{0.8}{P-value} & \scalebox{0.8}{***} & \scalebox{0.8}{**} & & \scalebox{0.8}{***} & ns &  \\
        \cline{2-9}
         & \multirow{2}{*}{Positional} & \scalebox{0.9}{Coefficients} & \textcolor{green}{- 0.6098} & \textcolor{green}{- 0.0629} & \textbf{9.69}  & \textcolor{green}{- 0.1571} & \textcolor{green}{- 0.0844} & \textbf{1.86} \\
         & & \scalebox{0.8}{P-value} & \scalebox{0.8}{***} & ns &  & \scalebox{0.8}{***} & \scalebox{0.8}{***} &\\
        \cline{2-9}
         & \multirow{2}{*}{Label} & \scalebox{0.9}{Coefficients} & \textcolor{green}{- 0.3572} & \textcolor{green}{- 0.0911} & \textbf{3.92} & \textcolor{green}{0.0602} & \textcolor{red}{0.0757} & 0.80 \\
         & & \scalebox{0.8}{P-value} & \scalebox{0.8}{***} & \scalebox{0.8}{**} & & \scalebox{0.8}{***} & \scalebox{0.8}{***} & \\
        \cline{2-9}
         & \multirow{2}{*}{Shape} & \scalebox{0.9}{Coefficients} & \textcolor{green}{- 0.1679} & \textcolor{green}{- 0.0707} & \textbf{2.37} & \textcolor{green}{- 0.0664} & \textcolor{green}{- 0.0081} & \textbf{8.20}  \\
         & & \scalebox{0.8}{P-value} & \scalebox{0.8}{***} & \scalebox{0.8}{***} & & \scalebox{0.8}{***} & \scalebox{0.8}{*} & \\
        \cline{2-9}
         & \multirow{2}{*}{Orientation} & \scalebox{0.9}{Coefficients} & \textcolor{green}{- 0.1746} & \textcolor{green}{- 0.0671} & \textbf{2.60} & \textcolor{green}{- 0.1073} & \textcolor{green}{- 0.0230} & \textbf{4.67}  \\
         & & \scalebox{0.8}{P-value} & \scalebox{0.8}{***} & \scalebox{0.8}{***} & & \scalebox{0.8}{***} & ns & \\
        \cline{2-9}
         & \multirow{2}{*}{Low-level Feature} & \scalebox{0.9}{Coefficients} & \textcolor{green}{- 0.1466} & \textcolor{green}{- 0.0457} & \textbf{3.21}  & \textcolor{green}{- 0.0493} & \textcolor{green}{- 0.0214} & \textbf{2.30} \\
         & &\scalebox{0.8}{P-value} & \scalebox{0.8}{***} & \scalebox{0.8}{**} & & \scalebox{0.8}{***} & \scalebox{0.8}{*} & \\
        
        \hline
        
         \multirow{12}{*}{CREPE} & \multirow{2}{*}{Phrasing} & \scalebox{0.9}{Coefficients} & \textcolor{green}{- 0.1149} & \textcolor{green}{- 0.0481} & \textbf{2.39} & \textcolor{green}{- 0.0025} & \textcolor{green}{- 0.0011} & \textbf{2.27} \\
         & & \scalebox{0.8}{P-value} & \scalebox{0.8}{***} & \scalebox{0.8}{***} & & ns & ns &  \\
        \cline{2-9}
         & \multirow{2}{*}{Positional} & \scalebox{0.9}{Coefficients} & \textcolor{green}{- 0.2914} & \textcolor{green}{- 0.1162} & \textbf{2.51} & \textcolor{red}{0.0192} & \textcolor{red}{0.0525} & 0.37 \\
         & & \scalebox{0.8}{P-value} & \scalebox{0.8}{***} & \scalebox{0.8}{***} & & ns & \scalebox{0.8}{**} & \\
        \cline{2-9}
         & \multirow{2}{*}{Label} & \scalebox{0.9}{Coefficients} & \textcolor{green}{- 0.1663} & \textcolor{green}{- 0.1147} & \textbf{1.45} & \textcolor{red}{0.0638} & \textcolor{red}{0.0407} & \textbf{1.57} \\
         & & \scalebox{0.8}{P-value} & \scalebox{0.8}{***} & \scalebox{0.8}{***} & & \scalebox{0.8}{***} & \scalebox{0.8}{***} &\\
        \cline{2-9}
         & \multirow{2}{*}{Shape} & \scalebox{0.9}{Coefficients} & \textcolor{green}{- 0.0952} & \textcolor{green}{- 0.0215} & \textbf{4.43} & \textcolor{green}{- 0.0196} & \textcolor{green}{- 0.0188} & \textbf{1.04} \\
         & & \scalebox{0.8}{P-value} & \scalebox{0.8}{***} & \scalebox{0.8}{*} & & \scalebox{0.8}{*} & \scalebox{0.8}{*} & \\
        \cline{2-9}
         & \multirow{2}{*}{Orientation} & \scalebox{0.9}{Coefficients} & \textcolor{green}{- 0.0797} & \textcolor{green}{- 0.0347} & \textbf{2.30} & \textcolor{green}{- 0.0320} & \textcolor{green}{- 0.0106} & \textbf{3.02} \\
         & & \scalebox{0.8}{P-value} & \scalebox{0.8}{***} & \scalebox{0.8}{**} & & \scalebox{0.8}{**} & ns & \\
        \cline{2-9}
         & \multirow{2}{*}{Low-level Feature} & \scalebox{0.9}{Coefficients} & \textcolor{green}{- 0.0919} & \textcolor{green}{- 0.0336} & \textbf{2.74} & \textcolor{green}{- 0.0202} & \textcolor{green}{- 0.0044} & \textbf{4.59} \\
         & &\scalebox{0.8}{P-value} & \scalebox{0.8}{***} & \scalebox{0.8}{**} & & \scalebox{0.8}{**} & ns & \\
        
        \hline
    \end{tabular}
    \caption{Both GPT-4o and Qwen2-VL exhibit greater bias impact at lower confidence levels, as reflected in absolute changes in both epistemic and aleatoric entropy with and without bias. This is supported by the consistent \textcolor{green}{negative} coefficients. Moreover, the bias impact on epistemic uncertainty correlates more strongly with confidence than on aleatoric uncertainty, as indicated by coefficient Ratio Epi./Ale.> 1 (\textbf{bolded}) and the relatively lower statistical significance of p-values for aleatoric entropy. (***p $\leq$ 0.001, **p $\leq$ 0.01, *p $\leq$ 0.05, ns=not significant p > 0.05)
    }
    \label{tab:empirical1}
\vspace{-1mm}
\end{table*}

When model confidence (self-perception) aligns with its true knowledge \citep{kadavath2022language, farquhar2024detecting}, it serves as a good estimate of the probability of correctness. As shown in Equation (\ref{eq:decompose}), the model's total uncertainty incorporates both epistemic uncertainty that indicates model confidence, and aleatoric uncertainty which does not. We use GPT-4o to evaluate the trade-off that bias mitigation introduces between these two types of uncertainty for uncertainty quantification.

\paragraph{Baselines.}
We focus on \textbf{Entropy} as our main baseline, given its strong performance in recent studies targeting closed-source LLMs \citep{kuhn2023semantic, farquhar2024detecting, yadkori2024believe}. We also include two commonly used baselines: the \textbf{Prob} (probability of the prediction) and the \textbf{\#Answers} (number of answers), as well as the recently proposed \textbf{Mutual Information} approach \citep{yadkori2024believe}, which adopts iterative prompting to estimate confidence based on the model's tendency to copy provided hints.


To address potential variation in token probabilities under identical decoding in closed-source models, we also introduce a \textbf{Repetitive-based} baseline that averages probabilities over multiple runs of the same prompt. This allows us to examine whether performance gains stem from better probability estimation simply through repeated sampling.

\paragraph{Analysis.} As shown in Table \ref{tab:model_perforamance_definite}, we observe that simple Repetitive-based samplings have minimal improvement over single-inference estimations.

\begin{table*}[h]
    \centering
    \scriptsize
    \begin{tabular}{ccccccccc}
        \toprule
        \multirow{2}{*}{\textbf{Dataset}} & \multirow{2}{*}{\textbf{Bias}} & \multirow{2}{*}{\textbf{Metrics}} & \multicolumn{3}{c}{\textbf{GPT-4o}} & \multicolumn{3}{c}{\textbf{Qwen2-VL}}\\
         \cmidrule(lr){4-6} \cmidrule(lr){7-9}
         & & &\textbf{Epistemic} & \textbf{Aleatoric}  & \textbf{Ratio Epi./Ale.} & \textbf{Epistemic} & \textbf{Aleatoric}  & \textbf{Ratio Epi./Ale.} \\
        \hline
        \multirow{12}{*}{VL\_Checklist} & \multirow{2}{*}{Phrasing} & \scalebox{0.9}{Coefficients} & \textcolor{green}{- 0.1651} & \textcolor{red}{0.0157} & \textbf{10.52} & \textcolor{green}{- 0.0158} & \textcolor{green}{- 0.0198} & 0.80 \\
        & & \scalebox{0.8}{P-value} & \scalebox{0.8}{***} & ns & & \scalebox{0.8}{*} & \scalebox{0.8}{*} &   \\
        \cline{2-9}
        & \multirow{2}{*}{Positional} & \scalebox{0.9}{Coefficients} & \textcolor{green}{- 0.7585} & \textcolor{green}{- 0.0499} & \textbf{15.2} & \textcolor{green}{- 0.1827} & \textcolor{green}{- 0.0722} & \textbf{2.53} \\
        & & \scalebox{0.8}{P-value} & \scalebox{0.8}{***} & ns & & \scalebox{0.8}{***} & \scalebox{0.8}{*} &  \\
        \cline{2-9}
        & \multirow{2}{*}{Label} & \scalebox{0.9}{Coefficients} & \textcolor{green}{- 0.3811} & \textcolor{green}{- 0.0898} & \textbf{4.24} & \textcolor{green}{-0.0338} & \textcolor{green}{-0.0233} & \textbf{1.45} \\
        & & \scalebox{0.8}{P-value} & \scalebox{0.8}{***} & \scalebox{0.8}{*} &  & ns & ns &  \\
        \cline{2-9}
        & \multirow{2}{*}{Shape} & \scalebox{0.9}{Coefficients} & \textcolor{green}{- 0.1542} & \textcolor{green}{- 0.0344} & \textbf{4.48}  & \textcolor{green}{- 0.0620} & \textcolor{green}{- 0.0013} & \textbf{47.69} \\
        & & \scalebox{0.8}{P-value} & \scalebox{0.8}{***} & ns & & \scalebox{0.8}{***} & ns &  \\
        \cline{2-9}
        & \multirow{2}{*}{Orientation} & \scalebox{0.9}{Coefficients} & \textcolor{green}{- 0.1441} & \textcolor{green}{- 0.0181} & \textbf{7.96} & \textcolor{green}{- 0.1309} & \textcolor{green}{- 0.0235} &  \textbf{5.57}  \\
        & & \scalebox{0.8}{P-value} & \scalebox{0.8}{***} & ns & & \scalebox{0.8}{***} & ns & \\
        \cline{2-9}
        & \multirow{2}{*}{Low-level Feature} & \scalebox{0.9}{Coefficients} & \textcolor{green}{- 0.1188} & \textcolor{green}{- 0.0121} & \textbf{9.82} & \textcolor{green}{- 0.0257} & \textcolor{green}{- 0.0011} & \textbf{23.36} \\
        & &\scalebox{0.8}{P-value} & \scalebox{0.8}{***} & ns & & \scalebox{0.8}{***} & ns &  \\
        
        \hline
        
        \multirow{12}{*}{CREPE} & \multirow{2}{*}{Phrasing} & \scalebox{0.9}{Coefficients} & \textcolor{green}{- 0.1019} & \textcolor{red}{ 0.0184} & \textbf{5.54}  & \textcolor{green}{- 0.0242} & \textcolor{red}{ 0.0097} & \textbf{2.49} \\
        & & \scalebox{0.8}{P-value} & \scalebox{0.8}{***} & ns & & \scalebox{0.8}{***} & ns & \\
        \cline{2-9}
        & \multirow{2}{*}{Positional} & \scalebox{0.9}{Coefficients} & \textcolor{green}{- 0.3929} & \textcolor{green}{- 0.0772} & \textbf{5.09} & \textcolor{green}{- 0.0951} & \textcolor{red}{0.0392} & \textbf{2.43} \\
        & & \scalebox{0.8}{P-value} & \scalebox{0.8}{***} & \scalebox{0.8}{*} & & \scalebox{0.8}{***} & ns &\\
        \cline{2-9}
        & \multirow{2}{*}{Label} & \scalebox{0.9}{Coefficients} & \textcolor{green}{- 0.2641} & \textcolor{green}{- 0.1082} & \textbf{2.44} & \textcolor{green}{- 0.0152} & \textcolor{red}{0.0184} & 0.83 \\
        & & \scalebox{0.8}{P-value} & \scalebox{0.8}{***} & \scalebox{0.8}{***} & & ns & ns & \\
        \cline{2-9}
        & \multirow{2}{*}{Shape} & \scalebox{0.9}{Coefficients} & \textcolor{green}{- 0.0580} & \textcolor{red}{0.0068} & \textbf{8.52} & \textcolor{green}{- 0.0147} & \textcolor{green}{- 0.0082} & \textbf{1.79}\\
        & & \scalebox{0.8}{P-value} & \scalebox{0.8}{***} & ns & & ns & ns & \\
        \cline{2-9}
        & \multirow{2}{*}{Orientation} & \scalebox{0.9}{Coefficients} & \textcolor{green}{- 0.0586} & \textcolor{green}{- 0.0206} & \textbf{2.84} & \textcolor{green}{- 0.0776} & \textcolor{green}{- 0.0095} & \textbf{8.17} \\
        & & \scalebox{0.8}{P-value} & \scalebox{0.8}{***} & ns & & \scalebox{0.8}{***} & ns &\\
        \cline{2-9}
        & \multirow{2}{*}{Low-level Feature} & \scalebox{0.9}{Coefficients} & \textcolor{green}{- 0.0741} & \textcolor{green}{- 0.0181} & \textbf{4.09} & \textcolor{green}{- 0.0152} & \textcolor{green}{- 0.0079} & \textbf{1.92} \\
        & &\scalebox{0.8}{P-value} & \scalebox{0.8}{***} & ns & & ns & ns &\\
        \hline
       
    \end{tabular}

    \caption{Both GPT-4o and Qwen2-VL exhibit greater bias-induced underestimation of epistemic uncertainty estimation when their confidence is lower, demonstrated by the \textcolor{green}{negative} coefficients and statistically significant p-values. In contrast, model confidence has no significant effect on the direction of bias effect in aleatoric entropy, supported by mostly insignificant p-values and mixed coefficient signs. The coefficient ratio Epi./Ale. > 1 is \textbf{bolded}. (***p $\leq$ 0.001, **p $\leq$ 0.01, *p $\leq$ 0.05, ns=not significant p > 0.05) }
    \label{tab:empirical2}
\vspace{-1mm}
\end{table*}

Bias mitigation consistently improves performance across all baselines. While no single bias mitigation method clearly outperforms the others, summing the entropy obtained from each bias removal leads to further performance gains. Similar accuracies across the ten bias-shuffled prompts shown in Appendix~\ref{sec:appendix accuracy} suggest that the improvement is not due to prompt quality differences.

Among bias mitigation strategies, combining three text-based methods yields the greatest performance improvement, increasing AUROC by 6.39\% on VL\_Checklist and 7.18\% on CREPE. In comparison, combining three image-based methods yields more modest improvement (2.07\% and 4.17\%, respectively), likely because image perturbation during training has already mitigated much of the image-based bias. Combining image- and text-based bias mitigation yields no further gains, suggesting text-based corrections capture most biases affecting uncertainty estimation. These findings highlight that bias removal is not only important for fairness but also critical for quantifying (epistemic) uncertainty when bias is significant.

The low performance of the Mutual Information method can be attributed to the concentration of its values as shown in Figure \ref{fig:model_curve_combined} in Appendix, a limitation shared by the \#Answers baseline. Specifically, the prevalence of identical Mutual Information values,  especially in low-uncertainty instances, limits its discriminative power and results in a low AUROC score. This makes it less suitable for high-stakes applications that demand a high abstention rate. In contrast, the text-based bias mitigation approaches remain robust across different thresholds.


\subsection{Relationship Between Model Confidence and Bias Impact}

We compute bias-free model confidence using the sum of the bias-free probabilities of correct options,  which serves as the independent variable. We then examine its relationship to absolute changes in measured epistemic and aleatoric entropy, comparing outputs with and without bias. Larger change indicates stronger bias impact. Results from two models and two datasets, as shown in Table \ref{tab:empirical1}, reveal consistent patterns across all biases:
\paragraph{Lower model confidence correlates with greater bias impact.} When the model exhibits lower bias-free confidence, its outputs tend to be more sensitive to bias, as evidenced by consistently negative coefficients for GPT-4o with only three exceptions in Qwen-2. 
\paragraph{Bias impact on epistemic uncertainty estimates is more strongly correlated with model confidence than on aleatoric uncertainty estimates.} This is evidenced by consistently higher coefficients for epistemic entropy compared to aleatoric entropy, as indicated by Ratio Epi./Ale. greater than one for GPT-4o, with only two exceptions for Qwen2-VL. In some cases, the bias impact on aleatoric uncertainty shows no significant correlation with bias-free model confidence, as indicated by large p-values (p > 0.05).

Similar results are obtained using bias-free epistemic entropy as the approximated model inconfidence, as shown in Appendix \ref{sec: appendix empirical}. The extended experiments on Claude-3.7-Sonnet with two randomly selected biases also align with these findings and are presented in the same appendix.

\subsection{Relationship Between Model Confidence and Bias-Induced Overconfidence}
\label{sec: emipirical2}

While lower model confidence leads to greater bias-induced changes, the direction of change is crucial. Greater under-confidence (i.e. overestimation of uncertainty) in lower bias-free confidence instances improves the robustness of estimated confidence ranking under estimation noise by amplifying the contrast between instances with low and high bias-free confidence. However, greater over-confidence in lower-confidence instances hurts the ranking performance of estimated confidence (see Figure \ref{fig:overconfidence}).

Therefore, we further examine how model confidence relates to bias impact on entropy reduction, subtracting measured entropy from a single prompt from that of bias-shuffled prompts. Results from two models and two datasets, as shown in Table \ref{tab:empirical2}, reveal consistent patterns across all biases:
\paragraph{Lower model confidence is associated with greater bias-induced underestimation of epistemic entropy (i.e., overconfidence).} When bias-free model confidence is lower, bias causes a larger reduction in epistemic entropy. This is evidenced by consistently negative coefficients for epistemic entropy reduction, with the majority of p-values indicating statistical significance.
\paragraph{Model confidence has no significant effect on the direction of bias-induced aleatoric entropy changes.} This is supported by the predominance of non-significant p-values and inconsistent coefficient signs for aleatoric entropy reduction.

Using bias-free epistemic entropy to approximate model inconfidence yields similar results, as shown in Appendix \ref{sec: appendix empirical}. The extended experiments on Claude-3.7-Sonnet with two randomly selected biases also align with these findings and are presented in the same appendix.

\section{Conclusion}

Removing three text-based biases and three image-based biases improves uncertainty quantification in the presence of aleatoric uncertainty, as measured by AUROC on GPT-4o. However, the improvement from image-based bias removal is smaller, likely due to existing image perturbation during training.

While uncertainty decomposes into epistemic and aleatoric components, our findings show that lower model confidence amplifies bias effects on measured uncertainties, with a greater amplification observed on epistemic than on aleatoric uncertainty. Moreover, while model confidence does not significantly affect the direction of bias-induced changes in measured aleatoric uncertainty, lower model confidence is associated with greater bias-induced underestimation of epistemic uncertainty (i.e. overconfidence).

Future work may leverage the distinct bias effects on these two types of uncertainty across varying confidence levels to develop more advanced techniques for disentangling them. In addition, our analysis could be extended to pure-text datasets once more challenging or less exposed multi-label benchmarks in LLMs pretraining become available.

\section*{Limitations}

\paragraph{Reliance on Token Probabilities.} While OpenAI provides token probabilities for its closed-source models, other LLMs impose stricter limitations. Some return only the predicted token’s probability without alternatives, while others, like Gemini, limit usage to one query per day. These constraints hinder the entropy-based uncertainty quantification method we use, which may require more samples to approximate the token probabilities.

\paragraph{Increase in Inference Cost.} While bias mitigation enhances the robustness of uncertainty quantification, it comes at the expense of the increased number of inferences. Shuffling prompts to account for each individual bias requires multiple model queries, increasing costs compared to single-inference methods.

\bibliography{custom}

\appendix

\clearpage

\section{Appendix}

\subsection{Mathematical Proof of Equation (\ref{eq:decompose})}
\label{sec:appendix proof}

\small
\noindent The entropy over the full distribution is
\begin{align}
\text{Entropy} &= -\sum_i P(i) \log P(i) \\
     &= -\sum_{i \in \text{correct}} P(i) \log P(i) - \sum_{i \notin \text{correct}} P(i) \log P(i)
\end{align}

\noindent The aleatoric entropy is defined as the entropy over the conditional distribution among correct options:
\begin{align}
\text{Aleatoric Entropy} &= - \sum_{i \in \text{correct}} \frac{P(i)}{P(\text{correct})} \log \left( \frac{P(i)}{P(\text{correct})} \right),
\end{align}

\noindent where $P(\text{correct}) = \sum_{i \in \text{correct}} P(i)$.\\

\noindent Multiply both sides by \( P(\text{correct}) \) to get:
\begin{align}
&\quad\:  P(\text{correct}) \cdot \text{Aleatoric Entropy} \notag \\
&= - \sum_{i \in \text{correct}} P(i) \log \left( \frac{P(i)}{P(\text{correct})} \right) \\
&= - \sum_{i \in \text{correct}} P(i) \log P(i) + \sum_{i \in \text{correct}} P(i) \log P(\text{correct}) \\
&= - \sum_{i \in \text{correct}} P(i) \log P(i) + P(\text{correct}) \log P(\text{correct}) 
\end{align}

\noindent Substitute this back into the total entropy:
\begin{align}
&\quad\: \text{Entropy} \notag \\
&= - \sum_{i \in \text{correct}} P(i) \log P(i) - \sum_{i \notin \text{correct}} P(i) \log P(i) \\
&= \left[ - \sum_{i \in \text{correct}} P(i) \log P(i) + P(\text{correct}) \log P(\text{correct}) \right] \\
&\quad - P(\text{correct}) \log P(\text{correct}) - \sum_{i \notin \text{correct}} P(i) \log P(i) \\
&= P(\text{correct}) \cdot \text{Aleatoric Entropy} + \\
&\quad \underbrace{ \left[ -P(\text{correct}) \log P(\text{correct}) - \sum_{i \notin \text{correct}} P(i) \log P(i) \right] }_{\text{Epistemic Entropy}} \\
&= P(\text{correct}) \cdot \text{Aleatoric Entropy} + \text{Epistemic Entropy}
\end{align}

\normalsize
\subsection{Details of Prompt Design}
\label{sec:appendix perturb}

Table \ref{tab:prompt exmaple} gives an example of vanilla prompt we used in our experiments.

\begin{table}[!ht]
\centering 
\small
\scalebox{0.85}{
\begin{tcolorbox}

\textbf{Prompt Example}

You are given an image and a set of descriptions. Your task is to evaluate each description and determine whether it is true based on the image.

Below are the descriptions:

\textcolor[rgb]{0,0,0.9}{0: person sitting in a boat with a paddle in the water. there is another paddle and boat in the water. the boat has writing on the side of it.}

\textcolor[rgb]{0,0,0.9}{1: person wearing shirt and captain on boat in water}

\textcolor[rgb]{0,0,0.9}{2: a boat with a paddle and captain on it, in dioxide}

\textcolor[rgb]{0,0,0.9}{3: captain of ground with yacht in water}

Provide one index of the descriptions that are true, regardless of the number of descriptions that you believe are true. Return your response as a single index without any additional explanations or text. Here is an example format for your response:

0

Use the provided format and structure for your response.

\end{tcolorbox}
}

\caption{The Vanilla Prompt example used to obtain greedy outputs.}
\label{tab:prompt exmaple}
\end{table}

\paragraph{Phrasing Bias.} We utilize GPT-4o to help paraphrase our default prompt shown in Table \ref{tab:prompt} while keeping the options unchanged. Table \ref{tab:prompts} lists all the rephrased prompts used in our experiments to perturb bias related to phrasing.

\paragraph{Positional Bias.} To perturb positional bias, we shuffle the assignments of option\_0, option\_2, option\_3, and option\_4 in the prompt template shown in Table \ref{tab:prompt}, while keeping the four labels in their natural order: 0, 1, 2, 3.

\paragraph{Label Bias.} To perturb label bias, we maintain the original positions of the options but shuffle the labels assigned to Label\_0, Label\_1, Label\_2, and Label\_3, such as 2, 0, 3, 1.

\paragraph{Shape Bias.} We resize images across different inputs by varying the length-to-width ratio from 0.5 to 1.5, intentionally distorting the shapes of objects in the images.

\paragraph{Orientation Bias.} We rotate images across different inputs by varying the rotated degrees from -45° to 45°. The rotation angles are kept relatively small to preserve the overall spatial relationships within the images.

\paragraph{Low-level Feature Bias.} We add random Gaussian noise with mean=0 and std=25 to the images across different inputs to disrupt local features while preserving their overall semantic meaning.

\subsection{Accuracy Comparison Between Default Prompt and Single Perturbed Prompt}
\label{sec:appendix accuracy}

\begin{table}[h]
    \centering
    \tiny
    \renewcommand{\arraystretch}{1.1}
    \begin{tabular}{|c|c|c|c|}
        \hline
        \textbf{Model} & \textbf{Dataset} & \textbf{Bias} & \textbf{Accuracy (\%)} \\
        \hline
        \multirow{12}{*}{GPT-4o} & \multirow{6}{*}{VL\_Checklist} & Default &  89.1  \\
        & & Phrasing & 86.5 \\
        & & Positional & 85.8  \\
        & & Label & 83.6 \\
        & & Shape & 87.5 \\
        & & Orientation & 86.5\\
        & & Low-level Feature & 86.7  \\
        
        \cline{2-4}
        
        & \multirow{6}{*}{CREPE} & Default & 73.3\\
        & & Phrasing & 73.7 \\
        & & Positional & 71.7 \\
        & & Label & 70.7 \\
        & & Shape & 73.1 \\
        & & Orientation & 72.9 \\
        & & Low-level Feature & 72.8 \\
        \hline
        
        \multirow{12}{*}{Qwen2-VL} & \multirow{6}{*}{VL\_Checklist} & Default &  92.1  \\
        & & Phrasing & 82.1 \\
        & & Positional & 82.8  \\
        & & Label & 77.9 \\
        & & Shape & 82.2 \\
        & & Orientation & 81.4\\
        & & Low-level Feature & 81.5 \\
        
        \cline{2-4}
        
        & \multirow{6}{*}{CREPE} & Default & 78.7\\
        & & Phrasing & 78.5 \\
        & & Positional & 78.7  \\
        & & Label & 77.9 \\
        & & Shape & 76.7 \\
        & & Orientation & 75.6\\
        & & Low-level Feature & 74.9 \\
        \hline
    \end{tabular}

    \caption{This table presents the accuracy achieved by the default prompt and the average accuracy achieved by each perturbed prompt with regard to each bias.}
    \label{tab:accuracy comparison}
\end{table}

Table \ref{tab:accuracy comparison} presents the accuracy comparison between the default prompt with greedy generation and each single bias-perturbed prompt used in our sampling method. The ranking of prompt performance does not correlate with their effectiveness in uncertainty quantification, indicating that the improvements in uncertainty quantification cannot be attributed to prompt quality.

\subsection{Details of Uncertainty Quantification Performance}
\label{sec: appendix uq}

\begin{figure*}[ht]
    \centering
    \begin{subfigure}[b]{0.43\linewidth}
        \centering
        \includegraphics[width=\linewidth]{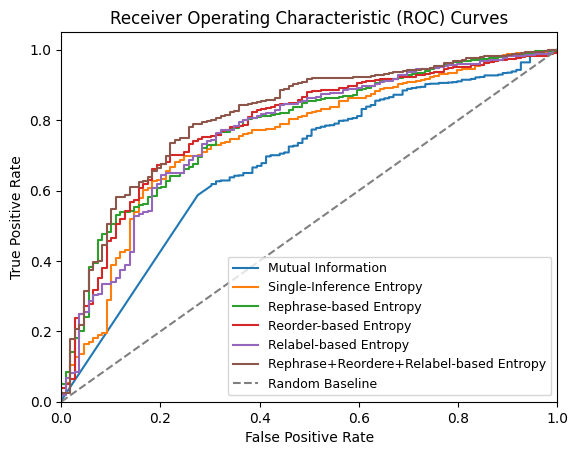}
        \caption{\small ROC curves on VL\_Checklist.}
        \label{fig:model_curve_vl}
    \end{subfigure}
    \hfill
    \begin{subfigure}[b]{0.43\linewidth}
        \centering
        \includegraphics[width=\linewidth]{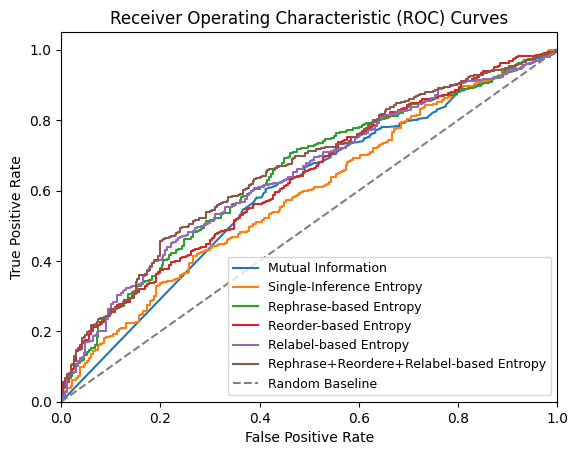}
        \caption{\small ROC curves on CREPE.}
        \label{fig:model_curve_crepe}
    \end{subfigure}

    \caption{Comparison of ROC curves for the text-based bias mitigation methods and baselines on two datasets using GPT-4o. The high prevalence of identical Mutual Information estimates makes it less suitable when a high abstention rate is required. The bias mitigation approach maintains robustness across different thresholds.}
    \label{fig:model_curve_combined}
\vspace{-0.3cm}
\end{figure*}

Figure \ref{fig:model_curve_combined} shows the ROC curves for text-based bias mitigation and baselines, providing more details of their performance across different threshold regions.

\subsection{More Empirical Results}
\label{sec: appendix empirical}

\begin{table}[h]
    \centering
    \begin{tabular}{|c|c|c|}
        \hline
        \textbf{Dataset} & \textbf{GPT-4o} & \textbf{Qwen2-VL}  \\
        \hline
        VL\_Checklist & 1.01 & 1.06 \\
        CREPE & 1.27 & 1.22 \\
        \hline
    \end{tabular}

    \caption{This table presents the ratio of Epistemic entropy to Aleatoric entropy across both datasets and models using the default prompt. Ratios closer to one indicate that aleatoric entropy is comparable in magnitude to epistemic entropy.}
    \label{tab:ratio comparison}
\end{table}

\paragraph{Relative Magnitudes of Epistemic and Aleatoric Entropy.} Table \ref{tab:ratio comparison} shows that the magnitude of aleatoric entropy is comparable to that of epistemic entropy.

\paragraph{Bias-free Epistemic Entropy as Model Confidence} We further validate our empirical findings by using the epistemic entropy after bias reduction, calculated from the average probabilities of ten shuffled prompts, as an approximation of the underlying model confidence. The results remain consistent with those obtained when approximating model confidence using the sum of the probabilities of correct options from the average probabilities. 

More specifically, the effects of bias, measured by changes in measured uncertainties, are more pronounced when model confidence is lower; in other words, when debiased epistemic entropy is higher. This is evidenced by consistently positive and statistically significant coefficients for bias-induced changes in measured epistemic uncertainty on GPT-4o. Qwen2-VL follows the same pattern, with exceptions for Label bias. For aleatoric uncertainty, GPT-4o also shows predominantly positive coefficients, whereas Qwen2-VL exhibits inconsistent coefficient directions with much smaller values, as indicated by Epi./Ale. ratios greater than one with the same two exceptions, and nonsignificant p-values. These results are detailed in Table \ref{tab:appendix empirical1}.

Lower model confidence is more strongly associated with greater underestimation of measured epistemic uncertainty, whereas it has no significant effect on the direction of bias-induced changes in measured aleatoric uncertainty. This is supported by the consistently positive and largely significant coefficients for the decrease in measured epistemic uncertainty, while the coefficients for the decrease in measured aleatoric uncertainty are predominantly insignificant except for the same two Qwen2-VL cases.

\paragraph{Extension to Claude-3.7-Sonnet.} Due to budget constraints, we randomly selected two biases for extended experiments on Claude-3.7-Sonnet with CREPE. Table~\ref{tab:claude-results} presents the results, which align with the findings observed on other models. More specifically, stronger bias effects on epistemic uncertainty estimation are associated with lower bias-free model confidence, as shown by negative and statistically significant coefficients. In contrast, the relationship for aleatoric uncertainty is weaker, as indicated by non-significant p-values and coefficients with mixed signs.

\begin{table*}[h]
    \centering
    \tiny
    \renewcommand{\arraystretch}{1.1}
    \begin{tabular}{cccccccc}
        \toprule
          \multirow{2}{*}{\textbf{Bias}} & \multirow{2}{*}{\textbf{Metrics}} & \multicolumn{3}{c}{\textbf{Absolute Change}} & \multicolumn{3}{c}{\textbf{Underestimation}}\\
         \cmidrule(lr){3-5} \cmidrule(lr){6-8}
         & &\textbf{Epistemic} & \textbf{Aleatoric}  & \textbf{Ratio Epi./Ale.} & \textbf{Epistemic} & \textbf{Aleatoric}  & \textbf{Ratio Epi./Ale.} \\
        \hline
          \multirow{2}{*}{Phrasing} & \scalebox{0.9}{Coefficients} & \textcolor{green}{-0.1579} & \textcolor{red}{0.0111} & \textbf{14.23}  & \textcolor{green}{-0.1238} & \textcolor{green}{- -0.0184} & \textbf{6.73}  \\
         & \scalebox{0.8}{P-value} & \scalebox{0.8}{***} & ns & & \scalebox{0.8}{***} & ns &  \\
        \hline
         \multirow{2}{*}{Label} & \scalebox{0.9}{Coefficients} & \textcolor{green}{-0.5420} & \textcolor{red}{0.2012} & \textbf{2.69} & \textcolor{green}{-0.5681} & \textcolor{red}{0.2094} & \textbf{2.71} \\
         & \scalebox{0.8}{P-value} & \scalebox{0.8}{***} & \scalebox{0.8}{**} & & \scalebox{0.8}{***} & \scalebox{0.8}{**} &  \\
    \end{tabular}

    \caption{This table reports the impact of phrasing and label bias on CREPE with Claude-3.7-Sonnet. Stronger bias effects on epistemic uncertainty estimation correspond to lower bias-free model confidence, as indicated by the negative coefficients with significant p-values. For aleatoric uncertainty, this relationship is weaker, reflected in the non-significant p-values and mixed coefficient signs.}
    \label{tab:claude-results}
\end{table*}

\begin{table*}[h]
    \centering
    \tiny
    \renewcommand{\arraystretch}{1.1}
    \begin{tabular}{ccccccccc}
        \toprule
         \multirow{2}{*}{\textbf{Dataset}} & \multirow{2}{*}{\textbf{Bias}} & \multirow{2}{*}{\textbf{Metrics}} & \multicolumn{3}{c}{\textbf{GPT-4o}} & \multicolumn{3}{c}{\textbf{Qwen2-VL}}\\
         \cmidrule(lr){4-6} \cmidrule(lr){7-9}
         & & &\textbf{Epistemic} & \textbf{Aleatoric}  & \textbf{Ratio Epi./Ale.} & \textbf{Epistemic} & \textbf{Aleatoric}  & \textbf{Ratio Epi./Ale.} \\
        \hline
          \multirow{12}{*}{VL\_Checklist} & \multirow{2}{*}{Phrasing} & \scalebox{0.9}{Coefficients} & \textcolor{green}{0.2622} & \textcolor{green}{0.0739} & \textbf{3.55}  & \textcolor{green}{0.0347} & \textcolor{red}{- 0.0056} & \textbf{6.20}  \\
         & & \scalebox{0.8}{P-value} & \scalebox{0.8}{***} & \scalebox{0.8}{***} & & \scalebox{0.8}{***} & ns &  \\
        \cline{2-9}
         & \multirow{2}{*}{Positional} & \scalebox{0.9}{Coefficients} & \textcolor{green}{0.4719} & \textcolor{green}{0.0379} & \textbf{12.45} & \textcolor{green}{0.1326} & \textcolor{red}{- 0.0654} & \textbf{2.03} \\
         & & \scalebox{0.8}{P-value} & \scalebox{0.8}{***} & \scalebox{0.8}{ns} &  & \scalebox{0.8}{***} & \scalebox{0.8}{***} &\\
        \cline{2-9}
         & \multirow{2}{*}{Label} & \scalebox{0.9}{Coefficients} & \textcolor{green}{0.2999} & \textcolor{green}{0.0575} & \textbf{5.22} & \textcolor{red}{-0.0255} & \textcolor{red}{-0.0828} & 0.31 \\
         & & \scalebox{0.8}{P-value} & \scalebox{0.8}{***} & \scalebox{0.8}{**} & & ns & \scalebox{0.8}{***} & \\
        \cline{2-9}
         & \multirow{2}{*}{Shape} & \scalebox{0.9}{Coefficients} & \textcolor{green}{0.2023} & \textcolor{green}{0.0822} & \textbf{2.46} & \textcolor{green}{0.0644} & \textcolor{green}{0.0144} & \textbf{4.47} \\
         & & \scalebox{0.8}{P-value} & \scalebox{0.8}{***} & \scalebox{0.8}{***} & & \scalebox{0.8}{***} & ns & \\
        \cline{2-9}
         & \multirow{2}{*}{Orientation} & \scalebox{0.9}{Coefficients} & \textcolor{green}{0.2126} & \textcolor{green}{0.0876} & \textbf{2.43} & \textcolor{green}{0.0916} & \textcolor{green}{0.0316} & \textbf{2.90} \\
         & & \scalebox{0.8}{P-value} & \scalebox{0.8}{***} & \scalebox{0.8}{***} & & \scalebox{0.8}{***} & \scalebox{0.8}{**} & \\
        \cline{2-9}
         & \multirow{2}{*}{Low-level Feature} & \scalebox{0.9}{Coefficients} & \textcolor{green}{0.1851} & \textcolor{green}{0.0536} & \textbf{3.45} & \textcolor{green}{0.0476} & \textcolor{green}{0.0205} & \textbf{2.32} \\
         & &\scalebox{0.8}{P-value} & \scalebox{0.8}{***} & \scalebox{0.8}{***} & & \scalebox{0.8}{***} & \scalebox{0.8}{**} & \\
        
        \hline
        
         \multirow{12}{*}{CREPE} & \multirow{2}{*}{Phrasing} & \scalebox{0.9}{Coefficients} & \textcolor{green}{0.1825} & \textcolor{green}{0.0558} & \textbf{3.27} & \textcolor{green}{0.0067} & \textcolor{red}{- 0.0020} & \textbf{3.30} \\
         & & \scalebox{0.8}{P-value} & \scalebox{0.8}{***} & \scalebox{0.8}{***} & & \scalebox{0.8}{*} & ns &  \\
        \cline{2-9}
         & \multirow{2}{*}{Positional} & \scalebox{0.9}{Coefficients} & \textcolor{green}{0.3344} & \textcolor{green}{0.0476} & \textbf{7.03} & \textcolor{green}{0.0139} & \textcolor{red}{-0.0508} & 0.27 \\
         & & \scalebox{0.8}{P-value} & \scalebox{0.8}{***} & \scalebox{0.8}{*} & & ns & \scalebox{0.8}{***} & \\
        \cline{2-9}
         & \multirow{2}{*}{Label} & \scalebox{0.9}{Coefficients} & \textcolor{green}{0.2129} & \textcolor{green}{0.0721} & \textbf{2.95} & \textcolor{red}{- 0.0744} & \textcolor{red}{- 0.0676} & \textbf{1.10} \\
         & & \scalebox{0.8}{P-value} & \scalebox{0.8}{***} & \scalebox{0.8}{***} & & \scalebox{0.8}{***} & \scalebox{0.8}{***} &\\
        \cline{2-9}
         & \multirow{2}{*}{Shape} & \scalebox{0.9}{Coefficients} & \textcolor{green}{0.1694} & \textcolor{green}{0.0423} & \textbf{4.00} & \textcolor{green}{0.0173} & \textcolor{red}{- 0.0029} & \textbf{5.97} \\
         & & \scalebox{0.8}{P-value} & \scalebox{0.8}{***} & \scalebox{0.8}{***} & & \scalebox{0.8}{*} & ns & \\
        \cline{2-9}
         & \multirow{2}{*}{Orientation} & \scalebox{0.9}{Coefficients} & \textcolor{green}{0.1723} & \textcolor{green}{0.0689} & \textbf{2.50} & \textcolor{green}{0.0227} & \textcolor{red}{- 0.0084} & \textbf{2.70}  \\
         & & \scalebox{0.8}{P-value} & \scalebox{0.8}{***} & \scalebox{0.8}{***} & & \scalebox{0.8}{*} & \scalebox{0.8}{ns} & \\
        \cline{2-9}
         & \multirow{2}{*}{Low-level Feature} & \scalebox{0.9}{Coefficients} & \textcolor{green}{0.1565} & \textcolor{green}{0.0517} & \textbf{3.03} & \textcolor{green}{0.0184} & \textcolor{green}{0.0064} & \textbf{2.88} \\
         & &\scalebox{0.8}{P-value} & \scalebox{0.8}{***} & \scalebox{0.8}{***} & & \scalebox{0.8}{***} & ns & \\
        
        \hline
    \end{tabular}
    \caption{Both GPT-4o and Qwen2-VL exhibit greater bias impact at lower confidence levels, as reflected in absolute changes in both epistemic and aleatoric entropy with and without bias. This is supported by the consistent \textcolor{green}{positive} coefficients. Moreover, the bias impact on epistemic uncertainty correlates more strongly with confidence than on aleatoric uncertainty, as indicated by coefficient Ratio Epi./Ale.> 1 (\textbf{bolded}) and the relatively lower statistical significance of p-values for aleatoric entropy. (***p $\leq$ 0.001, **p $\leq$ 0.01, *p $\leq$ 0.05, ns=not significant p > 0.05)
    }
    \label{tab:appendix empirical1}
\end{table*}

\begin{table*}[h]
    \centering
    \tiny
    \renewcommand{\arraystretch}{1.1}
    \begin{tabular}{cccccccccc}
        \toprule
        \multirow{2}{*}{\textbf{Dataset}} & \multirow{2}{*}{\textbf{Bias}} & \multirow{2}{*}{\textbf{Metrics}} & \multicolumn{3}{c}{\textbf{GPT-4o}} & \multicolumn{3}{c}{\textbf{Qwen2-VL}}\\
         \cmidrule(lr){4-6} \cmidrule(lr){7-9}
         & & &\textbf{Epistemic} & \textbf{Aleatoric}  & \textbf{Ratio Epi./Ale.} & \textbf{Epistemic} & \textbf{Aleatoric}  & \textbf{Ratio Epi./Ale.} \\
        \hline
        \multirow{12}{*}{VL\_Checklist} & \multirow{2}{*}{Phrasing} & \scalebox{0.9}{Coefficients} & \textcolor{green}{0.1537} & \textcolor{green}{0.0187} & \textbf{8.22} & \textcolor{green}{0.0230} & \textcolor{red}{- 0.0071} & \textbf{3.24} \\
        & & \scalebox{0.8}{P-value} & \scalebox{0.8}{***} & ns & & \scalebox{0.8}{***} & \scalebox{0.8}{ns} &   \\
        \cline{2-9}
        & \multirow{2}{*}{Positional} & \scalebox{0.9}{Coefficients} & \textcolor{green}{0.4874} & \textcolor{green}{0.0330} & \textbf{14.8} & \textcolor{green}{0.1311} & \textcolor{red}{- 0.0449} & \textbf{2.92} \\
        & & \scalebox{0.8}{P-value} & \scalebox{0.8}{***} & ns & & \scalebox{0.8}{***} & \scalebox{0.8}{*} &  \\
        \cline{2-9}
        & \multirow{2}{*}{Label} & \scalebox{0.9}{Coefficients} & \textcolor{green}{0.2942} & \textcolor{green}{0.0486} & \textbf{6.05} & \textcolor{green}{0.0267} & \textcolor{red}{-0.0070} & \textbf{3.81} \\
        & & \scalebox{0.8}{P-value} & \scalebox{0.8}{***} & ns &  & ns & ns &  \\
        \cline{2-9}
        & \multirow{2}{*}{Shape} & \scalebox{0.9}{Coefficients} & \textcolor{green}{0.1277} & \textcolor{green}{0.0438} & \textbf{2.92} & \textcolor{green}{0.0387} & \textcolor{red}{- 0.0033} & \textbf{47.69}  \\
        & & \scalebox{0.8}{P-value} & \scalebox{0.8}{***} & \scalebox{0.8}{*} & & \scalebox{0.8}{***} & ns &  \\
        \cline{2-9}
        & \multirow{2}{*}{Orientation} & \scalebox{0.9}{Coefficients} & \textcolor{green}{ 0.1590} & \textcolor{green}{0.0289} & \textbf{5.50} & \textcolor{green}{0.0883} & \textcolor{green}{0.0108} &  \textbf{8.18}  \\
        & & \scalebox{0.8}{P-value} & \scalebox{0.8}{***} & ns & & \scalebox{0.8}{***} & ns & \\
        \cline{2-9}
        & \multirow{2}{*}{Low-level Feature} & \scalebox{0.9}{Coefficients} & \textcolor{green}{0.1219} & \textcolor{green}{0.0192} & \textbf{6.35} & \textcolor{green}{0.0272} & \textcolor{red}{- 0.0080} & \textbf{3.4} \\
        & &\scalebox{0.8}{P-value} & \scalebox{0.8}{***} & ns & & \scalebox{0.8}{***} & ns &  \\
        
        \hline
        
        \multirow{12}{*}{CREPE} & \multirow{2}{*}{Phrasing} & \scalebox{0.9}{Coefficients} & \textcolor{green}{0.1577} & \textcolor{red}{- 0.008} & \textbf{197.13}  & \textcolor{green}{0.0116} & \textcolor{green}{0.0070} & \textbf{1.66} \\
        & & \scalebox{0.8}{P-value} & \scalebox{0.8}{***} & ns & & \scalebox{0.8}{*} & ns & \\
        \cline{2-9}
        & \multirow{2}{*}{Positional} & \scalebox{0.9}{Coefficients} & \textcolor{green}{0.4043} & \textcolor{green}{0.0327} & \textbf{12.36} & \textcolor{green}{0.0975} & \textcolor{red}{- 0.0433} & \textbf{2.25} \\
        & & \scalebox{0.8}{P-value} & \scalebox{0.8}{***} & ns & & \scalebox{0.8}{***} & \scalebox{0.8}{*} &\\
        \cline{2-9}
        & \multirow{2}{*}{Label} & \scalebox{0.9}{Coefficients} & \textcolor{green}{0.2890} & \textcolor{green}{0.0863} & \textbf{3.35} & \textcolor{green}{0.0171} & \textcolor{red}{- 0.0419} & 0.41 \\
        & & \scalebox{0.8}{P-value} & \scalebox{0.8}{***} & \scalebox{0.8}{***} & & ns & \scalebox{0.8}{**} & \\
        \cline{2-9}
        & \multirow{2}{*}{Shape} & \scalebox{0.9}{Coefficients} & \textcolor{green}{0.1425} & \textcolor{green}{0.0108} & \textbf{13.19} & \textcolor{green}{0.0282} & \textcolor{red}{- 0.0060} & \textbf{4.70} \\
        & & \scalebox{0.8}{P-value} & \scalebox{0.8}{***} & ns & & \scalebox{0.8}{**} & ns & \\
        \cline{2-9}
        & \multirow{2}{*}{Orientation} & \scalebox{0.9}{Coefficients} & \textcolor{green}{0.1478} & \textcolor{green}{0.0579} & \textbf{2.55} & \textcolor{green}{0.0738} & \textcolor{red}{- 0.0022} & \textbf{33.55} \\
        & & \scalebox{0.8}{P-value} & \scalebox{0.8}{***} & \scalebox{0.8}{***} & & \scalebox{0.8}{***} & ns &\\
        \cline{2-9}
        & \multirow{2}{*}{Low-level Feature} & \scalebox{0.9}{Coefficients} & \textcolor{green}{0.1299} & \textcolor{red}{- 0.0083} & \textbf{15.65} & \textcolor{green}{0.0186} & \textcolor{green}{0.0033} & \textbf{5.64} \\
        & &\scalebox{0.8}{P-value} & \scalebox{0.8}{***} & ns & & \scalebox{0.8}{**} & ns &\\
        \hline
    \end{tabular}

    \caption{Both GPT-4o and Qwen2-VL exhibit greater overconfidence in measured epistemic entropy due to bias when their confidence is lower, supported by \textcolor{green}{positive} coefficients and statistically significant p-values. In contrast, model confidence has no significant effect on the direction of aleatoric entropy changes caused by bias, as the directions of coefficients are inconsistent and p-values are not statistically significant. The coefficient ratio Epi./Ale. > 1 is \textbf{bolded}. (***p $\leq$ 0.001, **p $\leq$ 0.01, *p $\leq$ 0.05, ns=not significant p > 0.05)}
    \label{tab:appendix empirical2}
\end{table*}

\begin{table*}[!ht]
\centering 
\scriptsize
\resizebox{\textwidth}{!}{
\begin{tcolorbox}

\textbf{Prompt Template 1}

You are given an image and a set of descriptions. Your task is to evaluate each description and determine whether it is true based on the image.

Below are the descriptions:

\textcolor[rgb]{0,0,0.9}{\textless Options \textgreater}

Provide one index of the descriptions that are true, regardless of the number of descriptions that you believe are true. Return your response as a single index without any additional explanations or text. Here is an example format for your response:

0

Use the provided format and structure for your response.

\textbf{Prompt Template 2}

You are presented with an image and a list of descriptions. Your task is to assess each description and judge if it is true based on the image. 

The descriptions are listed below:

\textcolor[rgb]{0,0,0.9}{\textless Options \textgreater}

Indicate one index of the descriptions that are true, regardless of how many you think are correct. Return your response as a single index without any additional explanations or text. Here is an example format for your response:

0

Use the provided format and structure for your response.

\textbf{Prompt Template 3}

You have an image and several descriptions. Your task is to evaluate each description and determine its validity based on the image.

Below are the descriptions:

\textcolor[rgb]{0,0,0.9}{\textless Options \textgreater}

List one index of the descriptions that are true, even if multiple descriptions seem accurate. Return your response as a single index without any additional explanations or text. Here is an example format for your response:

0

Use the provided format and structure for your response.

\textbf{Prompt Template 4}

Given an image and a set of descriptions, your task is to evaluate each description and determine if it is true based on the image.

Here are the descriptions:

\textcolor[rgb]{0,0,0.9}{\textless Options \textgreater}

Provide one index of the descriptions that are true, even if multiple descriptions are accurate. Respond with a single index without any additional explanations or text. Here is an example format for your response:

0

Use the provided format and structure for your response.

\textbf{Prompt Template 5}

You have an image and a series of descriptions. Your task is to evaluate each description to determine its truthfulness based on the image.

Below are the descriptions:

\textcolor[rgb]{0,0,0.9}{\textless Options \textgreater}

Indicate one index of the true descriptions, even if there are multiple true descriptions. Return your response as a single index without any additional explanations or text. Here is an example format for your response:

0

Use the provided format and structure for your response.

\textbf{Prompt Template 6}

Given an image and several descriptions, your task is to evaluate each description and determine whether it is true based on the image.

Here are the descriptions:

\textcolor[rgb]{0,0,0.9}{\textless Options \textgreater}

Provide one index of the true descriptions, even if multiple descriptions are valid. Return your response as a single index without any additional explanations or text. Here is an example of how your response should look:

0

Use the provided format and structure for your response.

\textbf{Prompt Template 7}

You are provided with an image and a series of descriptions. Evaluate each description to determine if it is true based on the image.

Below are the descriptions:

\textcolor[rgb]{0,0,0.9}{\textless Options \textgreater}

Provide one index of the descriptions that are true, even if there are multiple descriptions that seem valid. Return your response as a single index without any additional explanations or text. Here is an example format for your response:

0

Use the provided format and structure for your response.

\textbf{Prompt Template 8}

Your task is to evaluate an image and a set of descriptions to determine if each description is true based on the image.

Here are the descriptions:

\textcolor[rgb]{0,0,0.9}{\textless Options \textgreater}

Provide an index of the true description(s), even if multiple descriptions seem correct. Return your response as a single index without any additional explanations or text. Here is an example format for your response:

0

Use the provided format and structure for your response.

\textbf{Prompt Template 9}

You have been given an image and a list of descriptions. Your task is to evaluate each description and determine if it is true based on the image.

The descriptions are as follows:

\textcolor[rgb]{0,0,0.9}{\textless Options \textgreater}

Provide one index of the descriptions that are true, even if you think more than one description is correct. Return your response as a single index without any additional explanations or text. Here is an example format for your response:

0

Use the provided format and structure for your response.

\textbf{Prompt Template 10}

You've been presented with an image alongside a series of descriptions. Your objective is to assess each description to determine its accuracy based on the image.

The descriptions are listed below:

\textcolor[rgb]{0,0,0.9}{\textless Options \textgreater}

You need to identify one description that is true, regardless of how many you think are correct. Please format your response as a single index without any additional explanations or text. Here is an example of how your response should look:

0

Ensure you adhere to this format and structure in your response..

\end{tcolorbox}
}

\caption{The ten prompts used to average the output distribution of Large Language Models in order to reduce phrasing bias through paraphrasing.}
\label{tab:prompts}
\end{table*}

\end{document}